# Human activity recognition based on time series analysis using U-Net

Yong Zhang [a*], Yu Zhang [a], Zhao Zhang [a], Jie Bao [a], Yunpeng Song [b]

[a]School of Electronic Engineering, Beijing University of Posts and Telecommunications, Beijing, 100876, China
[b]AIdong Super AI (Beijing) Co., Ltd, Beijing, 100007, China

**Abstract:** Traditional human activity recognition (HAR) based on time series adopts sliding window analysis method. This method faces the multi-class window problem which mistakenly labels different classes of sampling points within a window as a class. In this paper, a HAR algorithm based on U-Net is proposed to perform activity labeling and prediction at each sampling point. The activity data of the triaxial accelerometer is mapped into an image with the single pixel column and multi-channel which is input into the U-Net network for training and recognition. Our proposal can complete the pixel-level gesture recognition function. The method does not need manual feature extraction and can effectively identify short-term behaviors in long-term activity sequences. We collected the Sanitation dataset and tested the proposed scheme with four open data sets. The experimental results show that compared with Support Vector Machine (SVM), k-Nearest Neighbor (kNN), Decision Tree(DT), Quadratic Discriminant Analysis (QDA), Convolutional Neural Network (CNN) and Fully Convolutional Networks (FCN) methods, our proposal has the highest accuracy and F1-socre in each dataset, and has stable performance and high robustness. At the same time, after the U-Net has finished training, our proposal can achieve fast enough recognition speed.

## 1 Introduction

Human activity recognition (HAR) is the key technology of human-computer interaction and human activity analysis. The basic task of HAR is to select the appropriate sensor and deploy it to monitor and capture the user's activity [1]. HAR can be divided into two categories. The first one is video-based HAR, using video cameras to monitor the activity of the human body. Another is sensor-based HAR, which is based on time series data collected by sensors such as mobile phone built-in accelerometers [2-4], wrist-worn accelerometers [5-7], waist-mounted accelerometers [8-9], gyroscopes and magnetometers [10]. Due to the wide use of portable and wearable sensors with low cost, low power consumption, high capacity and miniaturization, HAR based on sensor data has become a research hotspot. The HAR system can be used in human-computer interaction application [11], behavior monitoring [12-13], health monitoring [14], smart home [15], medical care [16-17] and so on.

Data collected from portable and wearable sensors are usually time series data. Human activity recognition for time series is a complex process, which usually involves the following steps. First, preprocess the time series data such as smoothing, normalization [6], and separating gravity component [18] from acceleration data. Then segment the time series data, extract the feature of the data, and then classify by using the classification algorithm. It requires manual extraction of features to recognize human activity while using traditional machine learning methods. Domain features [19] includes mean, median, variance, standard variance, maximum, minimum, root mean square and etc. Frequency domain features [20] includes Fast Fourier Transform (FFT) coefficients and Discrete Cosine Transform(DCT) coefficients and etc. Considering that the extracted feature dimension is too high, Principal Cmponent Analysis (PCA) [21], Kernel Principal Component Analysis (KPCA) and Linear Discriminant Analysis(LDA) [22] are often used to reduce the dimension of the extracted feature to make it more robust. Support Vector Machine (SVM)[22-26], decision tree (DT) [19], K Nearest Neighbor (kNN) [27], naive Bayes [28], Hidden Markov Model (HMM) [29-30] are used to classify artificial extracted feature vectors. With the development of deep learning, the method of deep learning has been widely used in the study of HAR [31]. This method can automatically learn and extract features and omits the more complicated steps of manual feature extraction in traditional HAR research. Therefore, it is superior to the method of manual feature extraction. The convolution neural network is often used to analyze simple and complex activities from the time series data. The workload of feature engineering is greatly reduced by automatically learning and extracting features [6, 32].

*Corresponding author at: School of Electronic Engineering, Beijing University of Posts and Telecommunications, China.
E-mail address: yongzhang@bupt.edu.cn

Continuous segmentation of input sensor sequence data is a challenging task because the duration of human activity is different and the exact boundaries of activity are difficult to define. At present, both the manual feature extraction method and the automatic HAR method use windowing technology to divide the sensor signal into smaller time periods. And then the classification algorithm is used to recognize the activity in every window [33]. There are three different windowing techniques: fixed sliding window, event definition window and behavior definition window. Most researches on HAR adopt the method of fixed sliding window, which divides the signal into fixed-length windows and generates a label for all samples in one window. Compared with the other two methods, it is widely used because it does not need to preprocess the sensor signal. But using fixed sliding window to label data will also reduce the accuracy of HAR. All samples of a window may not always share the same label. The labelling method of fixed sliding window is used to generate a label for all the samples in the window. However, some samples will be mislabeled because the most-appearing class is select to represent the window[34]. This problem caused by fixed sliding window segmentation and labelling is called multi-classes window problem [35]. Multi-classes window problem is a common problem in HAR based on time series, which has a significant impact on the recognition of short-term activity sequences. It makes short-term activity classification challenging. In order to improve the recognition accuracy, the common method is to use a small-size window to segment [36], which is time-consuming by sacrificing speed in exchange for accuracy. At the same time, the small window may divide an action into several segments, which causes recognition errors, too.

In the field of deep learning in recent years, new neural network architectures has been developed to solve the problem of image semantic segmentation to realize the classification of image pixel levels, such as FCN [37], U-Net [38], SegNet [39], DeconvNet [40]. The most direct way to improve the recognition accuracy is to predict each sampling point of the time series. The multi-channel time series data can be regarded as a single pixel column and multi-channel image. This task can be regarded as a pixel level classification of the single pixel column images. Among them, U-Net network has been applied to the segmentation of audio time series to realize the separation of sound and background sounds [41]. In order to solve the multi-classes window problem caused by the fixed sliding window labelling, a U-Net based HAR algorithm is proposed in this paper. The input sensor time series is densely predicted, and the effective prediction of each sample label is realized.

The main contributions of this paper can be summarized as follows:

(1) To the best of our knowledge, U-Net network is applied to the HAR for the first time, which can predict the samples' label one by one. Furthermore, a large number of experiments verify the better prediction accuracy due to its unique network structure;

(2) The activity data of the triaxial accelerometer is mapped into an image with the single pixel column and multi-channel which is input into the U-Net network for training and recognition.

(3) A new dataset named Sanitation is released to test the HAR algorithm's performance, which will befit the researchers in this field.

The structure of the rest of the paper is as follows: Section 2 provides an overview of related work. Section 3 explains the basic principles of dense labelling and prediction. section 4 describes HAR algorithm based on U-Net is proposed in Section 4. The assessment of experiments and results is shown in section 5. The conclusion is given in section 6.

## 2 Related work

The early research of HAR is based on the manual feature extraction from time series data, and then classifies by various classification algorithms. The current machine learning algorithms for HAR can be divided into two categories: classification based on discriminative model and classification based on generative model [42].

The classification method of HAR based on discriminative model mainly includes SVM, DT, kNN and Artificial Neural Network (ANN). In [43], the authors extract the autoregressive coefficients of accelerometer data as the characteristics of activity recognition, and use SVM to classify human activity. The average recognition result of running, standing, jumping and walking is 92.25%. In [19], the author uses the built-in accelerometer to classify five activities and constructs a location-independent activity recognition model based on the DT algorithm. Khan A M et al establish a hierarchical scheme, in which the upper layer uses the autoregressive model of acceleration signal to generate the augmented feature vector and the lower layer processes the eigenvector of triaxial accelerometer data through LDA analysis and ANNs. The average recognition accuracy of 15 specific activities is 97.9% [44]. Preece S J et al use the nearest neighbor classifier to classify and analyze the daily rows based on acceleration time series. They adopt robust individual-based cross validation method. The classification accuracy on the best feature set reaches 95% [27].

The classification methods of HAR based on generative model mainly include HMM and naive Bayes method.

Lester J et al propose a dynamic behavior recognition method [29]. The recognition accuracy is 95% by capturing temporal regularity and smoothness by HMM. Lee S et al propose a HAR algorithm based on semi-Markov random domain with an average accuracy of 88.47% and 86.68%, higher than the result of HMM [45]. Long X et al use PCA to reduce the dimension of feature vector and Bayesian classifier to classify daily action. The accuracy rate is 80%, which is better than that of DT [28].

In recent years, depth learning theory has made great achievements in static image feature extraction, and has been gradually extended to the study of time series data. The deep learning methods for HAR can be summarized into three categories. The first category uses Convolutional Neural Network (CNN) to automatically extract features from sensor data for recognition. Song-Mi Lee et al propose a 1-dimensional CNN to identify human behavior for triaxial acceleration sensors collected by smart phones. The recognition accuracy of three simple postures is 92.71% [46]. In [34], the author constructs a HAR model based on CNN, and modifies the convolution kernel to adapt to the characteristics of triaxial acceleration signal. The average recognition accuracy is 93.8%. Panwar M et al investigate a depth learning framework for predicting the arm motion in daily behavior by using a hand-mounted triaxial accelerometer. The CNN is used to automatically extract useful features. The average recognition accuracy is 99.8%, better than clustering, linear discriminant analysis and SVM method [6]. The second category is to use Recurrent Neural Network (RNN) to capture the time dependence of sensor data. Guan Y and Ploetz propose a HAR model based on Ensembles of deep Long Short Term Memory, which has achieved a good recognition effect on Opportunity, PAMAP2 and Skoda datasets [47]. In [48], the author proposes a HAR model based on a Binarized Long Short-Term Memory Network (B-BLSTM-RNN). The third category is to use mixed model to identify human activity. Ordóñez and Roggen D propose a depth framework of HAR based on CNN and LSTM unit cyclic neural network. The accuracy of recognition on the Opportunity and Skoda dataset is higher than that on the previous report by 9%. It is suitable for multimodal wearable sensors and can accurately model the feature of real-time dynamic changes without using professional knowledge to design features [34]. NY Hammerla and Shane Halloran et al proposed a HAR scheme based on deep convolution and cyclic model for wearable device sensor data [49].

More recently, Rui Yao et al. proposed a human activity recognition algorithm based on full convolutional neural network[35], which realized the dense prediction of human activity sequences from wearables and conducted extensive experiments on three datasets. Their experiments obtain 88.7% with weighted F-measure on Opportunity Locomotion dataset, 59.6% on Opportunity Gesture dataset, 89.3% on subject 1 for Hand Gesture dataset, 88.3% on subject 2 for Hand Gesture dataset, and 79.0% on the self-collected Hospital dataset.

Different from the above work, the U-Net based on HAR algorithm predicts the labels of each sample in the input time series precisely so as to overcome the multi-classes window problem existing in the sliding window method. Its unique network structure makes the prediction more accurate than existing solutions.

## 3 Dense labelling and prediction

The traditional fixed sliding window is used to divide the continuous time series signals into fixed-length windows. All samples in the window are labeled with the same label by the window labelling strategy. This labelling method based on fixed sliding window can lead to multi-class window problem. The classifier outputs incorrect context information, resulting in lower recognition accuracy. As shown in Fig. 1, there are two common sliding window labelling strategies [34]. The one is to select the most frequent sample class in the window as the label of the window. Another is to select the sample class of the last time step in the window as the label of the window. Both the two methods lead to the incorrect labelling, and reducing the recognition accuracy. In order to solve the multi-class window problem caused by fixed sliding window, the dense labelling is proposed [35]. Dense labelling provides the correct label information for each sample and improves the recognition accuracy of classifier.

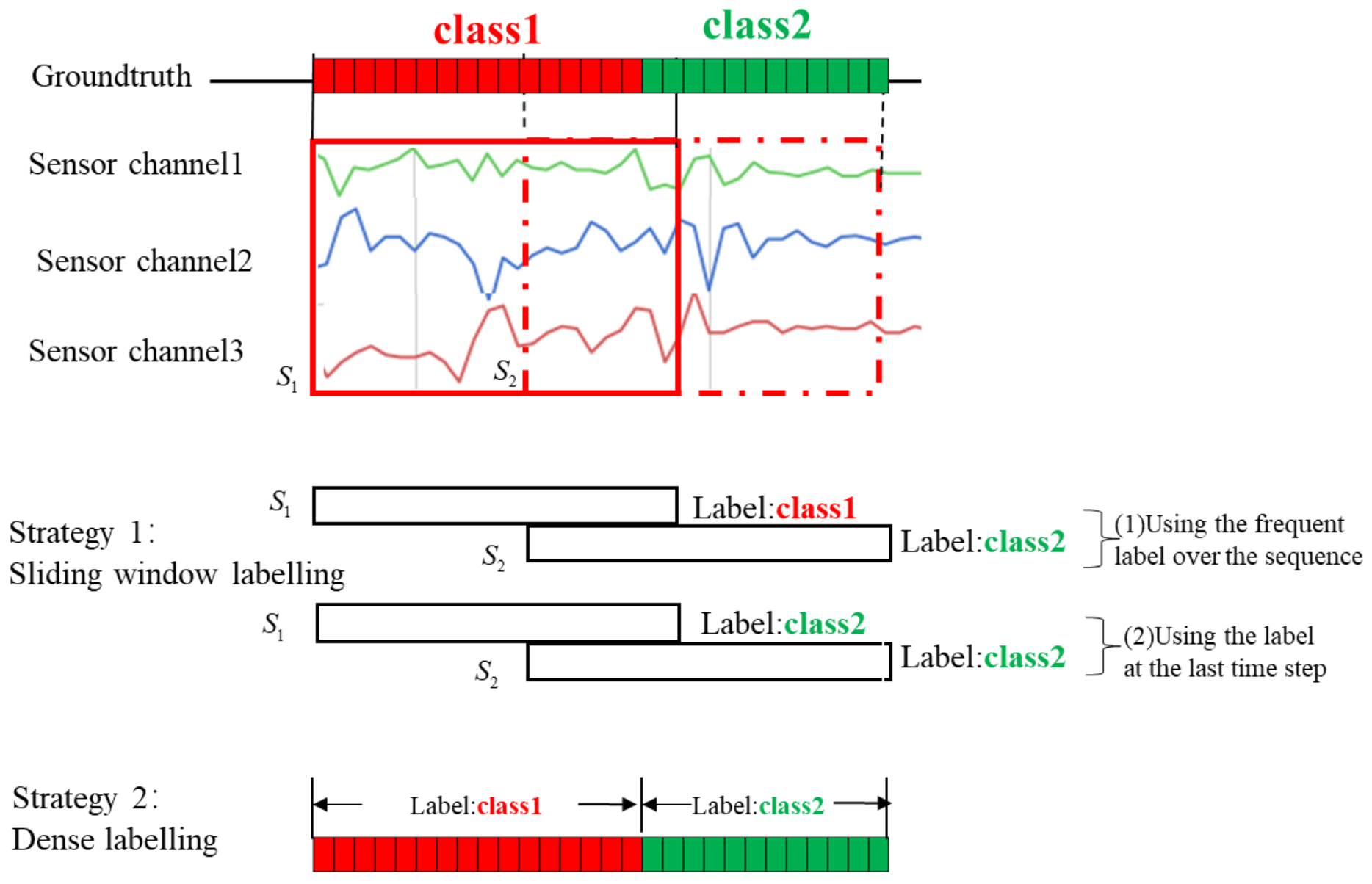

Figure 1. Sliding window labelling and dense labelling

Figure 1 describes the two different sequence labelling methods: sliding window labelling and dense labelling. Red solid line frame represents window $S_1$ and red dashed line frame represents window $S_2$. According to sliding window labelling strategy 1, the most frequent sample class in the window $S_1$ is class 1, so it is labeled as class 1. Whereas virtually all samples of $S_1$ include class 1 and class 2 information. This results in the loss of class 2 label information and the learning of incorrect context information. In contrast, the dense labelling is used to label the classes of each sample in the sequence rather than window-based labelling. Thus, it preserves all the label information of the sequence.

The prediction of each sample label is called dense prediction. The traditional CNN architecture reduces the resolution of top-level output by introducing maximum pooling operation. For image target recognition task, this property can reduce the sensitivity of image shift in recognition task [49]. However, for the time series recognition task, using CNN architecture will result in a mismatch between the output prediction label length and the input prediction time series length. Therefore, the traditional CNN architecture limits the implementation of dense prediction. But it is suitable for sliding window prediction. The sliding window labeled data is input into the traditional CNN architecture and output a single prediction label for each segment of window data. U-Net network has been successfully developed to image semantic segmentation by introducing the operation of up-convolution to achieve the same resolution between the top output and input. The dense labeled window data is input into the U-Net architecture, which allows equal input and output length. That is to say, the output of each input sample has a corresponding predictive label. U-Net architecture can be implemented as Fig. 2(b).

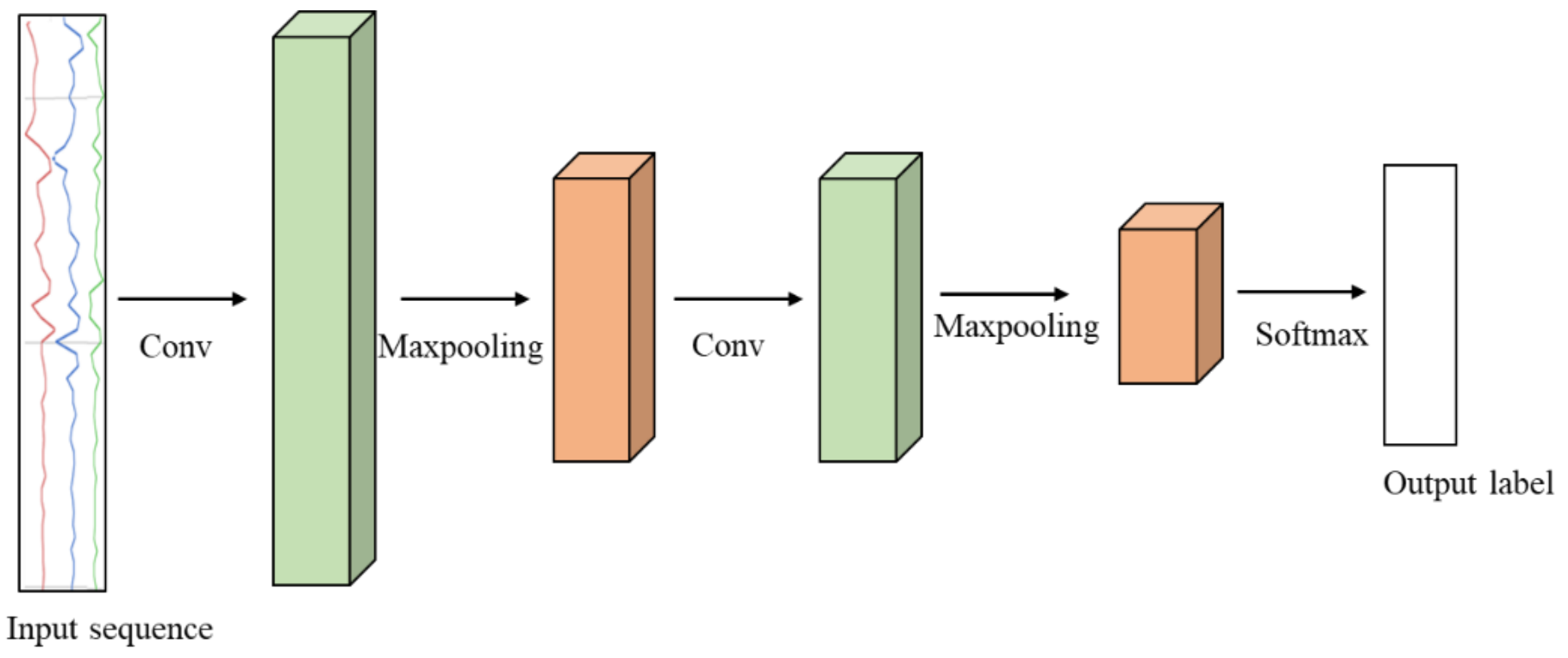

(a) Sliding window prediction of CNN

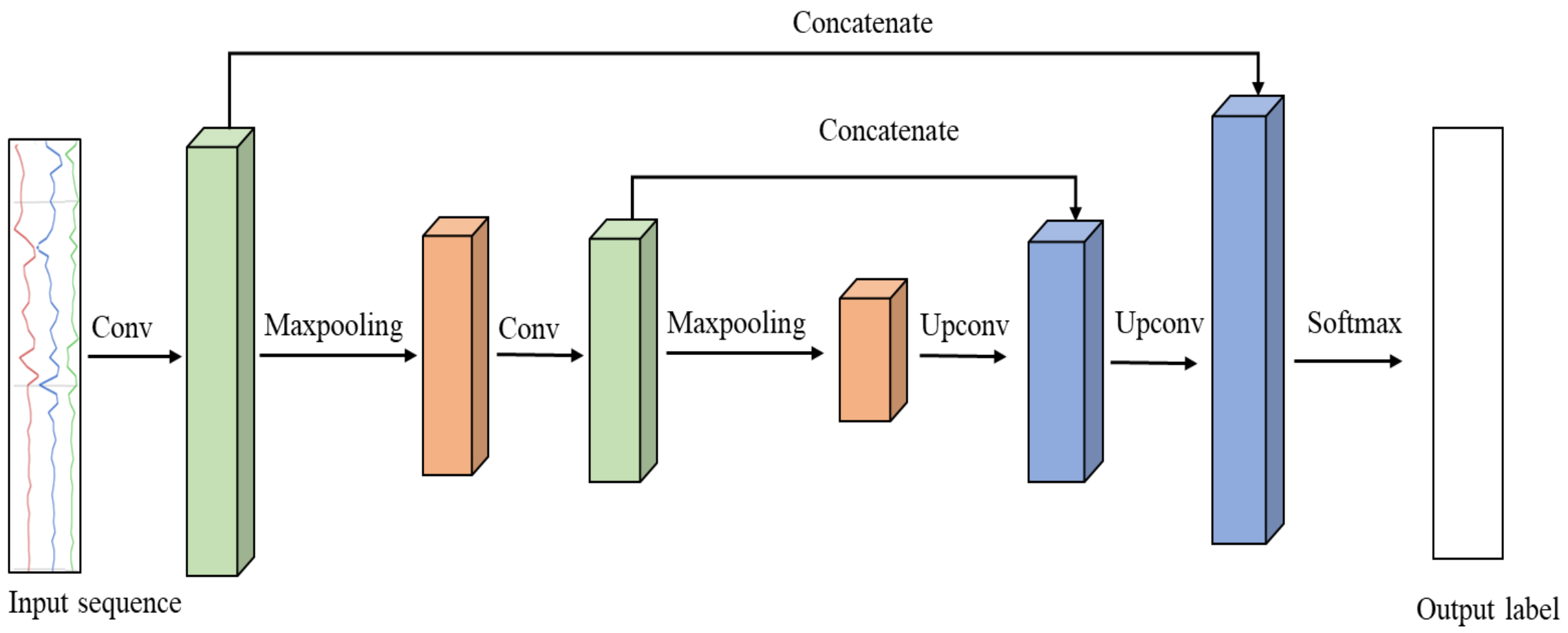

(b) Dense prediction of U-Net

Figure 2. The different prediction method of CNN and U-Net

Figure 2 illustrates the different prediction forms of traditional CNN and U-Net networks. Figure 2 (a) describes a simple CNN. The resolution of the top-level output is lower than the resolution of the underlying input due to the maximum pooling operation in the network. While inputting the time series data to be predicted, the length of the output prediction label sequence is smaller than that of the input time series. CNN network can only predict the window label. Figure 2 (b) describes a simple U-Net network. The top-level output has the same resolution as the underlying input because of the up-convolution operation in the network. While inputting the time series data to be predicted, the length of the outputted prediction label sequence has the same length as the input time series. The U-Net network can complete the label prediction of each sampling point, that is, the dense prediction.

In conclusion, the dense labelling of time series by U-Net overcomes the multi-class window problem. And the dense prediction of the time series is realized by directly inputting the time series to U-Net network. This method provides an effective solution to the HAR problem using time series.

## 4 4. HAR algorithm based on U-Net

### 4.1 U-Net

U-Net is the development and extension on CNN that can make use of the multi-layer network structure to automatically learn the features. In CNN, shallow convolution layer learns local features, and deep convolution layer learns more abstract features. These features, on the one hand, enhance CNN's classification ability, on the other hand, lose the detailed information of objects. So it is impossible to classify each pixel.

In order to achieve pixel level classification, Olaf Ronneberger et al proposed an end-to-end U-Net network for biomedical image segmentation to improve the precise localization and segmentation of neuronal structures in microscopic images [12]. The purpose of U-Net is to find the corresponding class of each pixel through feature processing and to realize the classification of pixel level.

The neural network consists of two paths. One is a contracting path to capture contextual information on the left and the other is an expansive path, which is symmetric, like the uppercase letter U. Therefore, it is named U-Net. The contracting path on the left is the same as the traditional convolution network, including convolution, pooling, activation and so on, which is used to capture the content of the image. The expansive path on the right gradually improves the resolution of the network layer by up-sampling. By stacking the up-sampling network layer and the corresponding convolutional network layer on the left side, the shallow and deep network information can be merged to offset the loss of information caused by the previous pooling operation. Finally, when the resolution of network layer and input layer is the same, two convolution operations and one full connection operation are performed again.

The input feature map of each layer in the network can be regarded as a three-dimensional tensor of $h_l \times w_l \times f_l$ size. $h_l$ and $w_l$ represent the height and width of the input feature map of layer $l$. $f_l$ represents the number of the input feature map of layer $l$. If the network inputs an RGB image of three color channels with $h_1 \times w_1$ pixels, $f_1$=3. The input data vector at the location $(i, j)$ in a particular layer is denoted $x_{i,j}$. Then $y_{i,j}$ denotes the output vector of the layer. It can be calculated by the formula (1),

$$y_{i,j} = f(\{x_{i+i',j+j'}\}_{-\frac{k_i-1}{2} \le i' \le \frac{k_i-1}{2}, -\frac{k_j-1}{2} \le j' \le \frac{k_j-1}{2},}) \tag{1}$$

Where the size of the convolution kernel is represented by $k_i$ and $k_j$. $f(\bullet)$ represents the type of layer: matrix multiplication of convolution layer, maximum operation of maximum cell layer, nonlinear operation with activation function and etc. $y_{i,j}$ is the output of layer $l$ and the input of layer $l+1$. Its size can be expressed as $h_{l+1} \times w_{l+1} \times f_{l+1}$, among which $h_{l+1} = (h_l - k_i)/s_i + 1$ and $w_{l+1} = (w_l - k_j)/s_j + 1$. The stride of movement is represented by $s_i, s_j$. By performing proper filling operations on input feature maps, the output feature map can have the same resolution as the input feature map.

Considering the two-dimensional time series that we input, the first dimension is the sample sequence time dimension. The second dimension is the number of channels of the sensor, that is, the number of sensor data axes. So it can be regarded as a single pixel column and multi-channel image which is input into the U-Net network. Each sample point of the time series can be predicted by means of the prediction ability of the U-Net pixel level.

### 4.2 Network Architecture

The proposed network structure is shown in Fig. 3. The network inputs two-dimensional time series data collected by the sensor. It can be regarded as a single pixel column and multi-channel image, in which the size of the image is $(1, N, N_s)$. $N$ represents the sampling points of the input sub-sequence. $N_s$ represents the number of channels for the sensor. The network consists of 28 convolution layers, which are divided into two parts: contracting network and expansive network. Each layer of the contracting network is composed of two convolution layers with a size of $1\times3$ and a pooling layer with a size of $1\times2$. The size of the feature map remains the same by setting the appropriate filling after each convolution operation. And it is activated by the Restricted Linear Unit (ReLU) function. Then the size of the feature map is reduced by half after the pooling operation. The number of feature maps at each layer of the contracting network is constant, but the number of feature maps at the next level is twice that of the previous one. For example, the number of feature maps at the first layer of the contracting network is $f$, and the number of the second feature maps at the second level is $2f$. $f$ in our experiment is 32. Each layer of the expansive network corresponds to the level of the contracting network. Each layer of the expansive network is mainly composed of one up-convolution layer with a size of $1\times2$ and two convolution layers with s size of $1\times3$.The output feature map of the convolution operation doubles the size of the output feature map so that it has the same resolution as the output feature map of the corresponding contracting layer to achieve the merging. Then two convolution operations and the activation of the ReLU function are performed. Finally, by mapping each feature vector to the corresponding class through a convolutional network with a size of $1\times1$ and Softmax classifier, the dense prediction results of the input time sequence are obtained. The dimension is $(1, N, N_c)$, where $N_c$ is the number of human activity classes.

### 4.3 Network training

When using U-Net to predict HAR dense label, the infinitely long time series can not be considered as network input. The input time series have to be divided into overlapping long sub-sequences. The dense prediction is performed after the sub-sequence is input into U-Net network in order to realize the prediction of each sample point of the sub-sequence.

The whole time series is defined as $X$, and the length of the sub-sequence is defined as $N$. A large number of different sub-sequences are extracted from the whole time series by sampling and sliding mode. Thus, the continuous division of the sub-sequence to the whole time series is realized. It is worth noting that the sub-sequences here are different from traditional sliding windows. For sub-sequences, each sample needs in a sub-sequence to be predicted, that is, each sample corresponds to a label, while the sliding window is a single label for all samples in a window. The length of the sub-sequence can be set longer to speed up the training speed.

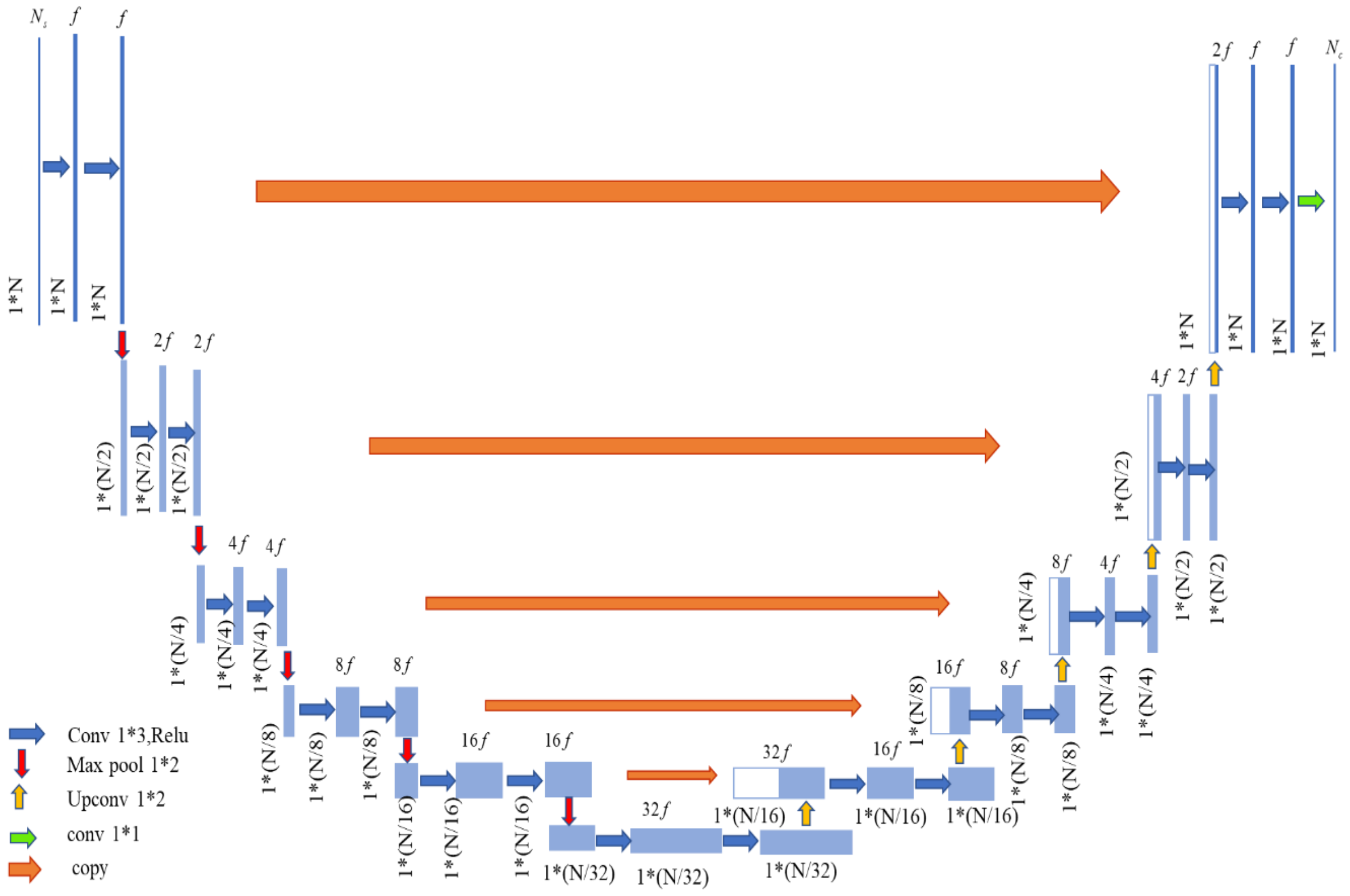

Figure 3. Network architecture of the U-Net used in this paper

In order to identify the boundaries of different activities and provide useful contextual information in the actual scene, the sub-sequences are generated by sequential sliding. Considering the overlap between adjacent sub-sequences, the prediction of sample labels in overlapping parts may produce ambiguity. So the overlap rate between sub-sequences is set to 0. When the starting point of the $t$ th sub-sequence is introduced as $p_t$, the $t$ th sub-sequence is expressed as,

$$x[1, p_t : p_t + N, N_s] \tag{2}$$

According to the set overlap rate, the starting point of the $t+1$ th sub-sequence is $p_{t+1}=p_t + N$.

Given a set of input sequences and labels, the goal of network training is to estimate the appropriate parameters $(W, b)$ of U-Net network to achieve accurate dense prediction. This is achieved by minimizing the loss values of all samples from each sub-sequence in the training dataset. Use the negative logarithmic likelihood function as the loss function:

$$l(x, y; W, b) = \sum_j^N l^{'}(x, y_j; W, b) \tag{3}$$

Where $l^{'}(x, y_j; W, b) = -\log(p(y_j \mid x, W, b))$ represents the loss function of the $j$ th sample in a sub-sequence.

It can be seen that a sub-sequence is a mini-batch, and the training of all the sub-sequences is an epoch. Each sub-sequence is also updated by each mini-batch through the Adam algorithm [51].

## 5 Experiment

This section focuses on the experiments of HAR using U-Net. Compared with other algorithms on different datasets, the experiments evaluate the effectiveness of applying U-Net to HAR. First, the datasets and parameter configuration used in the experiment are introduced. Then, the performance of each algorithm on each dataset is compared synthetically. Finally, a new evaluation index is proposed to suit the dense labelling scene.

### 5.1 Datasets

In this paper, we use five datasets, including WISDM dataset [2], UCI HAR dataset [52], UCI HAPT dataset [23], UCI OPPORTUNITY dataset [53] and the self-collected Sanitation dataset.

1. WISDM dataset

WISDM dataset is collected by the Wireless Sensor Data Mining laboratory (WISDM) and consists of two different versions, namely, the WISDM Activity Prediction dataset and the WISDM Actitrackers dataset. The former is adopted in this paper and referred as WISDM dataset in the following. In this paper, data of 36 subjects are collected by smart phone. The data are triaxial acceleration time series, namely X axis acceleration time series, Y axis acceleration time series and Z axis acceleration time series. The sampling frequency is 20Hz. It includes 1098209 samples, that is, the total acquisition time is about 915 minutes. The dataset consists of six types of activities: walking, jogging, upstairs, downstairs, sitting and standing. The proportion of activity samples is shown in Fig. 4 (a).

WISDM dataset not only provides the original acceleration time series but also provides the converted data. The original acceleration time series is regarded as an activity of 200 continuous sampling points every 10 seconds, and then converted into a $1 \times 43$ vector by calculating 43 eigenvalues. A total of 5418 eigenvectors are generated. In other words, the input matrix of $1098209 \times 3$ is converted to the characteristic matrix of $5418 \times 43$, and the data is transformed from input space to feature space.

2. UCI HAR dataset

UCI HAR dataset is one of the most famous open datasets in the field of HAR, provided by University of California Irvine. Wearing the smartphone at the waist and using the built-in accelerometer and gyroscope, 30 volunteers aged between 19 and 48 attend the data collection. The data collected include a 3-axis acceleration time series and a 3-axis angular velocity time series. The sampling frequency is 50 Hz. The dataset contains 6 types of activities, respectively: WALKING, WALKING_UPSTAIRS, WALKING_DOWNSTAIRS, SITTING, STANDING, LAYING. The proportion of activity samples is shown in Fig. 4 (b).

Preprocess the original time series. Specifically, use a low-pass filter to filter the noise firstly. After that, the time series is divided into a segment sub-sequence by a sliding window with a size of 2.56 seconds (including 128 sample points) and 50% overlap. The original time series is divided into 10299 sub-sequences. For each sub-sequence, 561 eigenvalues are calculated and converted into a $1 \times 561$ eigenvector. The dataset is randomly divided into two parts: 70% training set and 30% test set. The number of feature vectors contained in the training set and test set is

7352 and 2947 respectively.

3. UCI HAPT dataset

UCI HAPT dataset is an extension of the UCI HAR dataset. The data were also collected from 30 volunteers aged 19 to 48 using a smartphone worn around the waist. The datasets include six basic types of activity in the UCI HAR dataset as well as six postural transitioning activities. Specifically, it is stand-to-sit, sit-to-stand, sit-to-lie, lie-to-sit, stand-to-lie and lie-to-stand. The proportions of activity samples are shown in figure 4 (c).

The original time series are divided into 10929 sub-sequences and extracted features. The dataset is also divided into 70% training set and 30% test set. The number of eigenvectors contained in the training set and test set are 7767 and 3162 respectively. The UCI HAR dataset, unlike the UCI HAR dataset, also provides raw time series that are not processed by sliding windows.

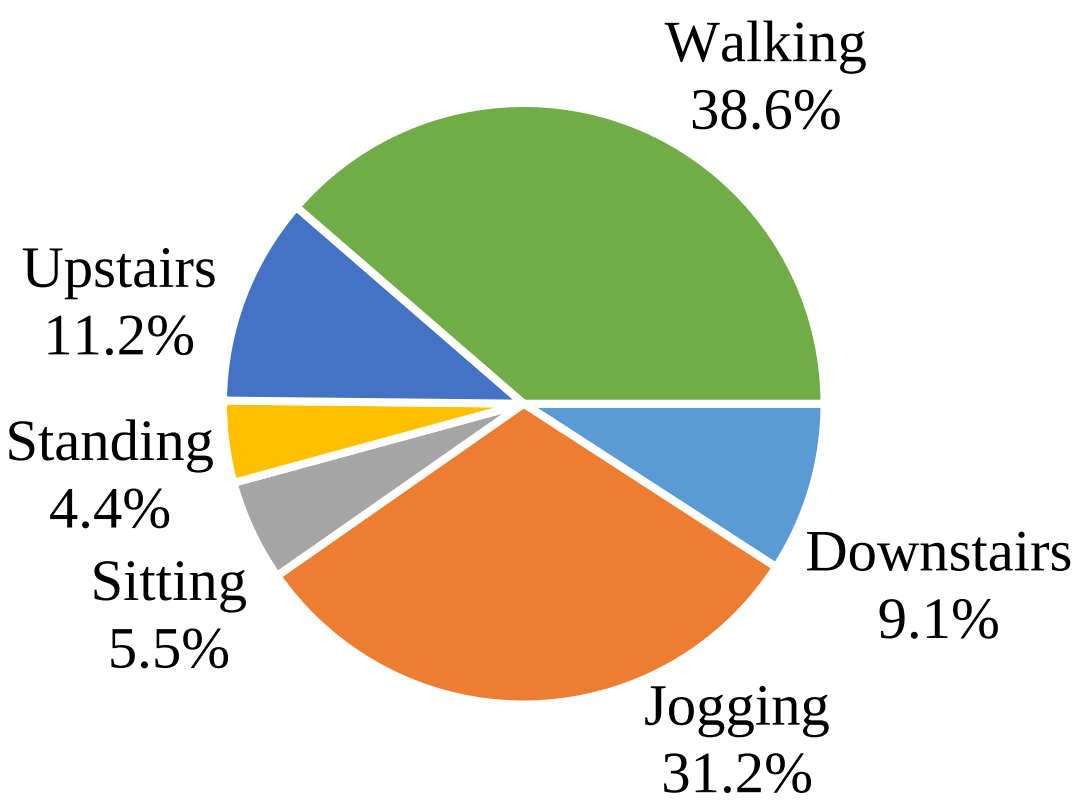

(a) WISDM dataset

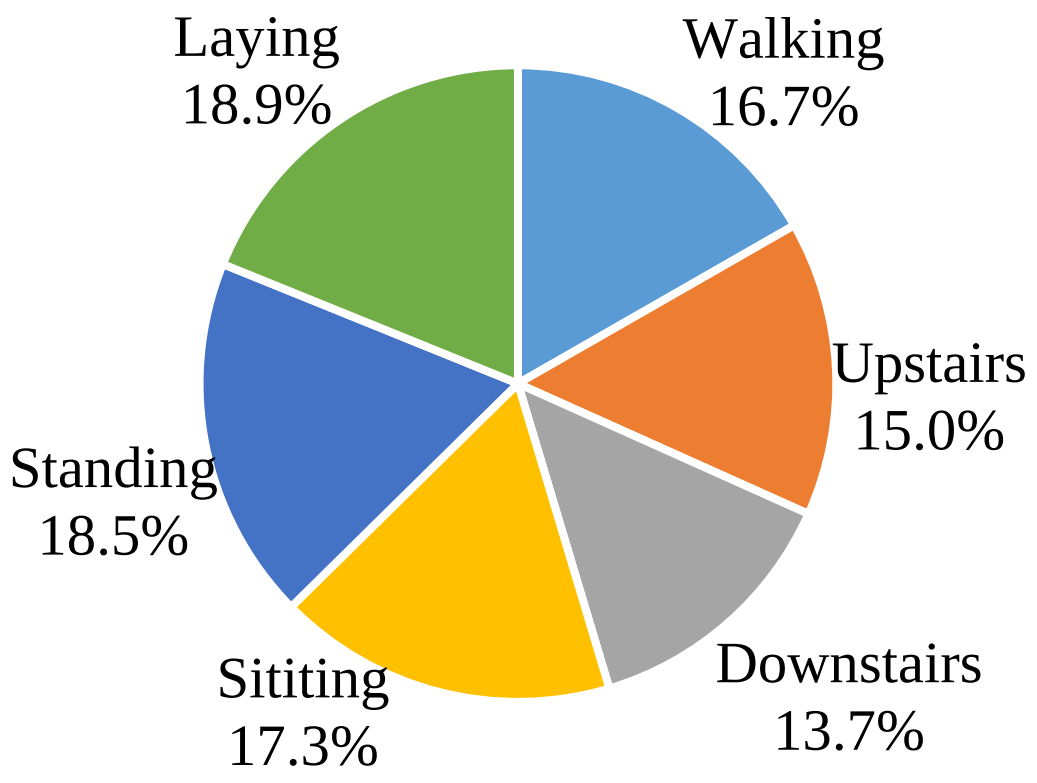

(b) UCI HAR dataset

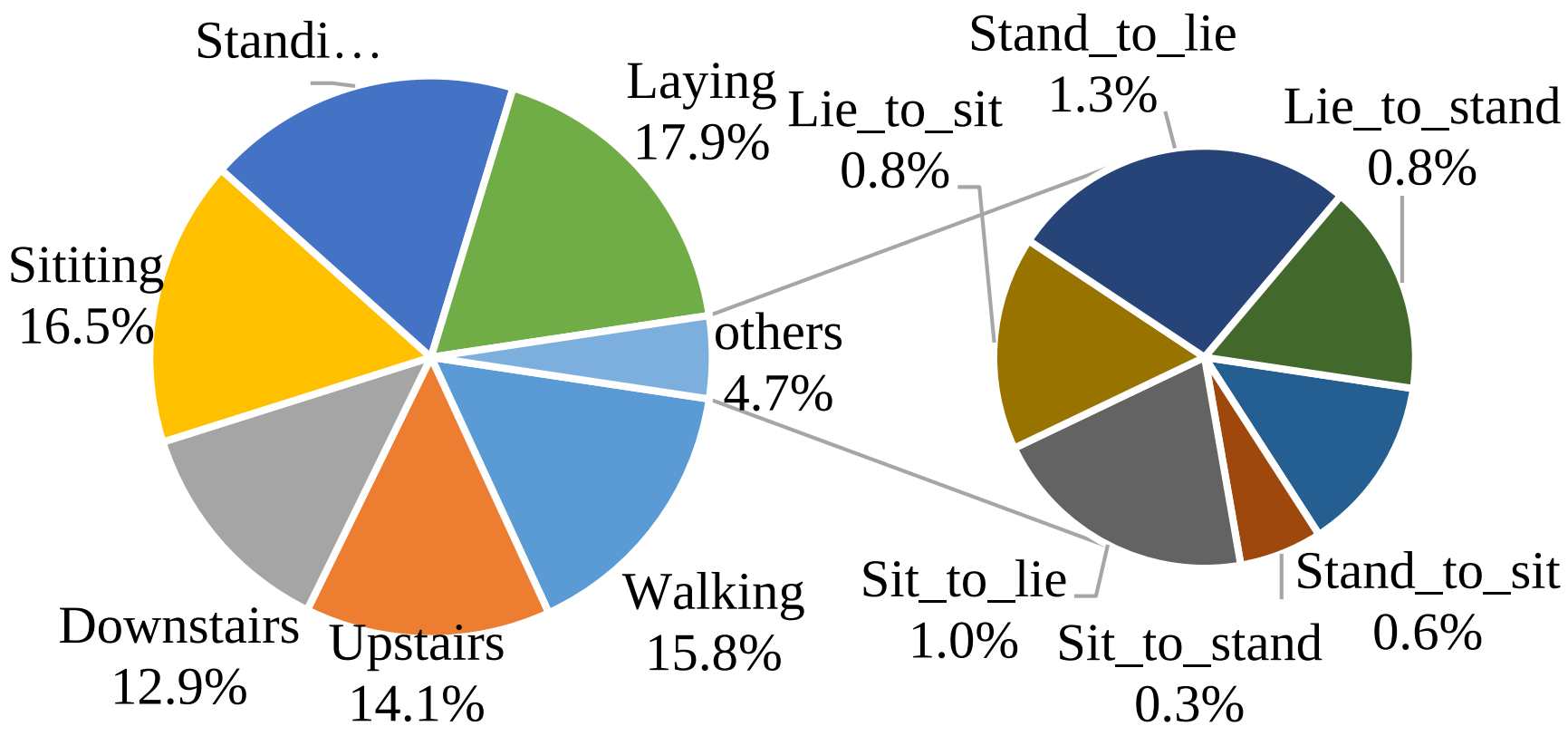

(c) UCI HAPT dataset

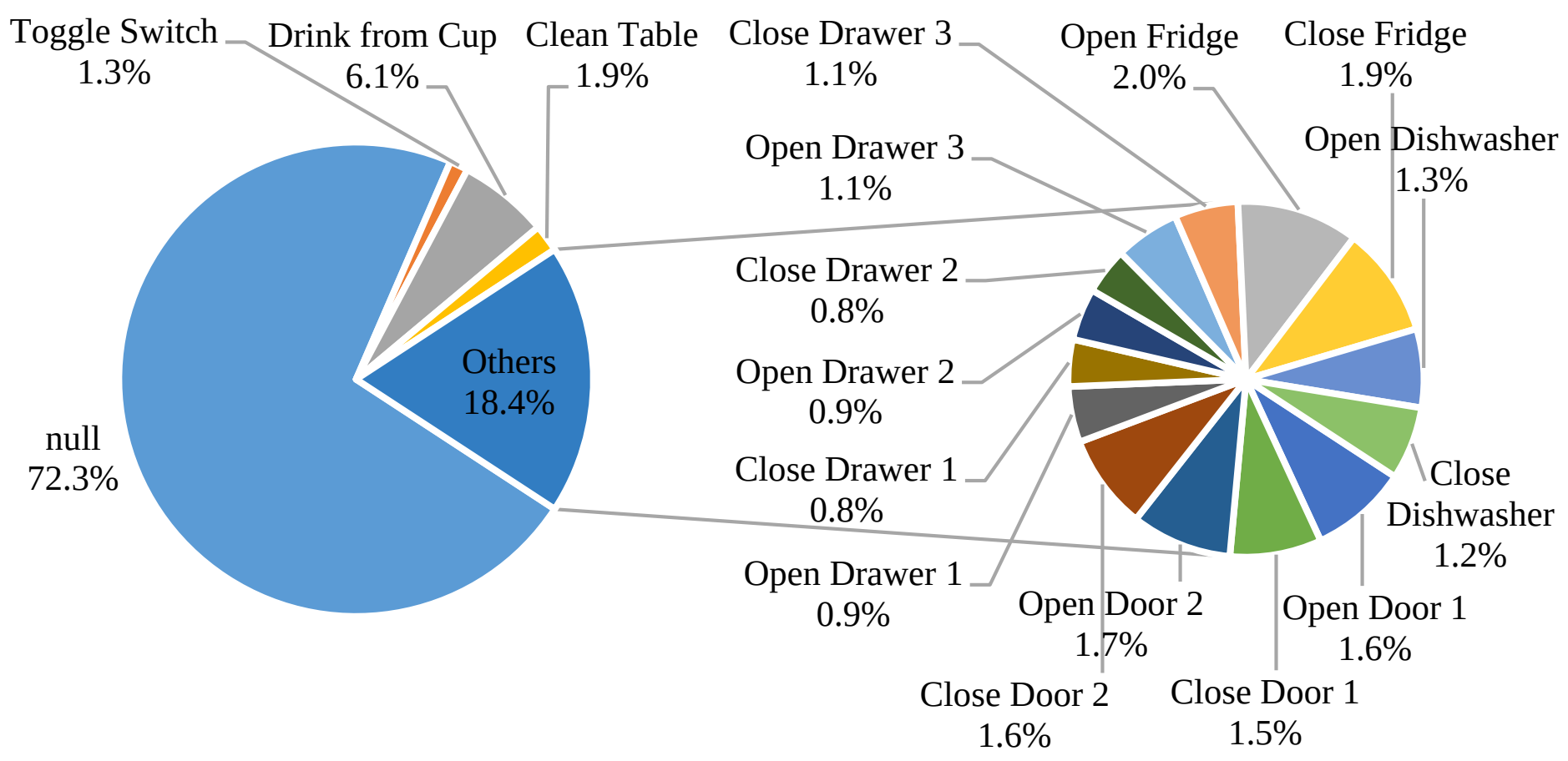

(d) UCI OPPORTUNITY Gesture dataset

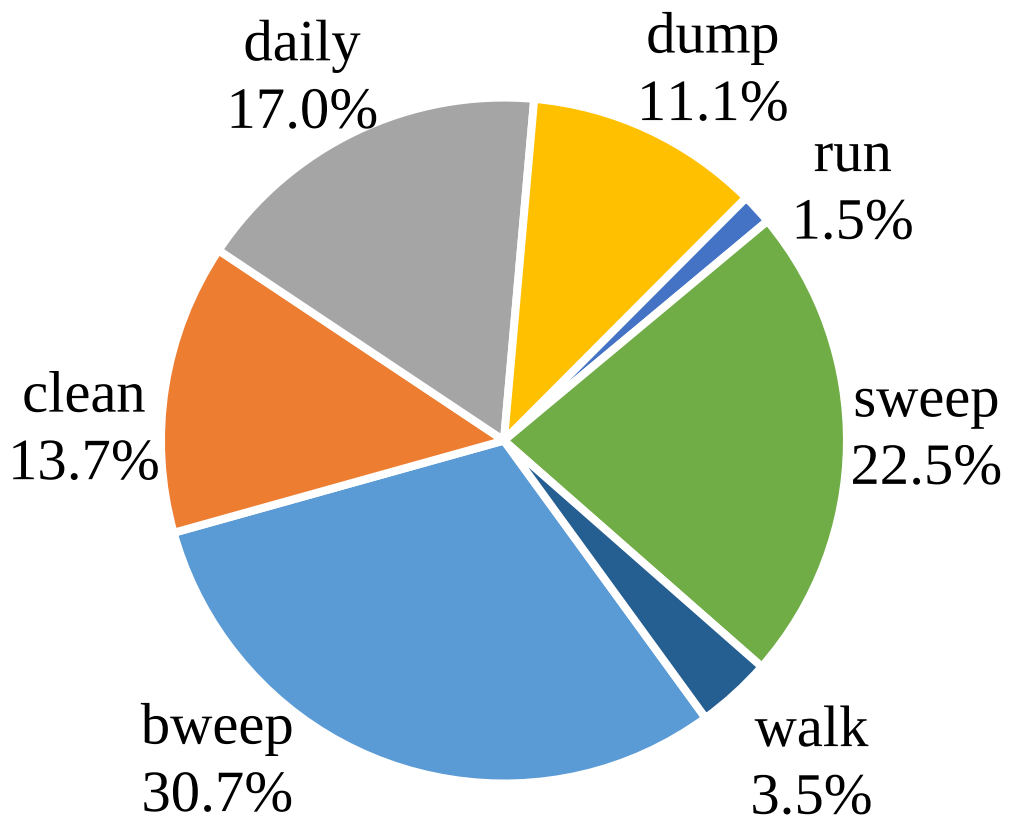

(e) Sanitation dataset

Figure 4. Percentage of activity of the five datasets

4. UCI OPPORTUNITY dataset

UCI OPPORTUNITY dataset is a benchmark dataset to verify HAR algorithm. The dataset is collected from a limited experimental environment. The sensors include body-worn sensor, object sensor, and ambient sensors. The sampling rate is 30 Hz. There are four individuals, each of whom conducts six rounds of experiments. The first five of them are Daily living activity, in which individuals perform their daily activities in a natural manner. The sixth round is conducted in the Drill mode. An individual performs an action in a prearranged order. The size of the dataset is 2551×242. That is, 2551 samples are included, and the number of attributes of each sample is 242. The dataset is not divided into the training set and the test set.

The activity collected can be divided into four categories: locomotion, low level action, middle level gesture (Gesture) and high level action. The Gesture dataset contains 17 gestures and a null class. The Locomotion dataset consists of four basic classes and one null class. The four basic classes are as follows: sit, stand, lie and walk. In this paper, the Gesture data is experimented. The proportions of various activity samples are shown in Fig. 4 (d).

5. Sanitation dataset

The self-collected Sanitation dataset is collected from the open environment. A triaxial accelerometer worn in a wrist smart watch is used to collect seven types of daily work activity data of sanitation workers. The sampling frequency is 25 Hz. These seven types of activity are: walk, run, sweep, bweep (sweep using big broom), clean, dump and daily activities. The size of the whole dataset is 266555 x 3, which contains 266555 samples. Each sample contains X, Y and Z three axis acceleration values. The proportion of various types of activity samples is shown in Fig. 4 (e). Fig. 5 is a snapshot the Sanitation dataset, where the horizontal axis represents the indexes the samples, and the longitudinal axis represents the values of the accelerometer, where the value 1000 represents 1g, that is, 9.81m/s$^2$.

The original time series consists of 5026 sub-sequences of different lengths, each representing a specific activity class. For each sub-sequence, 57 features, including time-domain and frequency-domain features, are extracted. And a 1×57 feature vector is generated. The input matrix of 266555×3 is transformed into 5026×57 feature matrix.

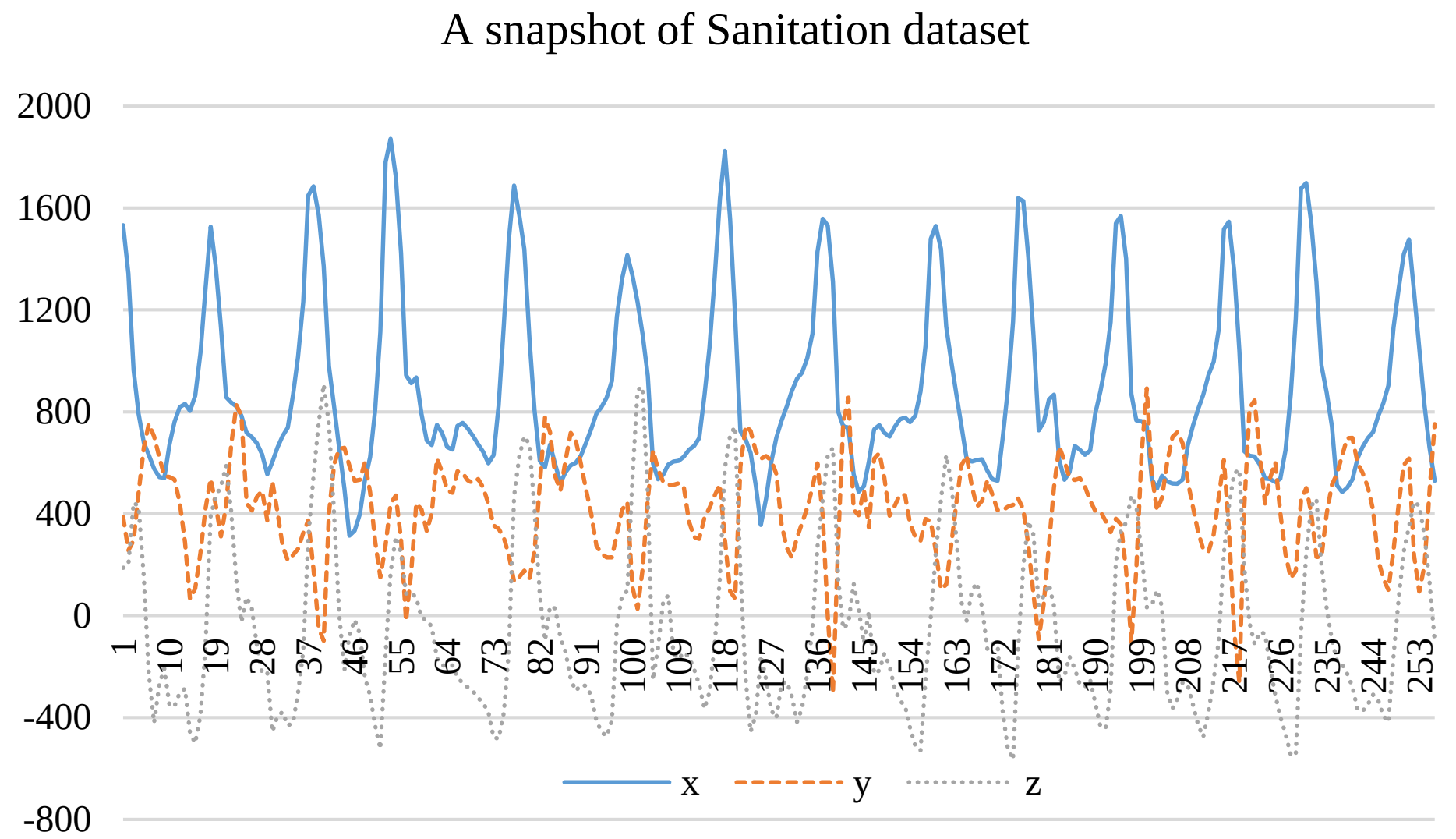

Figure 5. A snapshot of Sanitation dataset

## 5.2 Experiment configuration

In order to train U-Net network, and the learning rate is set to 0.001, the training batch size is set to 32 and the training batch (epoch) is set to 100. The network parameters are optimized and updated by the Adam algorithm. Each sub-sequence represents a mini-batch, and the length of the sub-sequence, that is, the number of sample points included, is set to 224. The overlap rate between sub-sequences is 0. All the sub-sequences are divided into training set and test set, in which the test set accounts for 30%.

In addition, in order to compare with the proposed U-Net method, we also test six other algorithms, namely SVM, kNN(k=5), DT, QDA, CNN. and FCN, which is proposed to be applied to dense labelling in [35]. For these algorithms, the ratio of the test set is set to 30%, and the parameters of the same algorithm are consistent on each dataset.

The hardware used in the experiment is a desktop computer equipped with an Intel (R) Core (TM) i5-4460 @ 3.2GHz CPU and an NVIDIA GeForce GTX 1060 6G GPU. The programming language used in the experiment is Python3.6.4.

### 5.3 Unified evaluation index

The algorithm based on sliding window divides the time series into segments of sub-windows and predict a label of each sub-window. The evaluation index such as the accuracy and F1 values is calculated by comparing the given label and prediction label for the sub-window. However, for dense prediction, what is predicted is not the label of the sub-window, but the label of each sample point in the whole time series. Therefore, when calculating the accuracy and F1 values, it is obtained by comparing the given label and the prediction label of each sample point in the whole time series.

Due to the difference of the evaluation index calculation method between the sliding window-based algorithm and U-Net, a unified model evaluation index is defined. The dense labelling evaluation index calculation method is adopted uniformly. The given label and prediction label of each sample point in the whole time series are compared and calculated. For a sliding window based algorithm, the label of the sub-window needs to be assigned to each sample point in the sub-window. Then all the sample points in the sub-window are connected together and the labels of each sample point in the whole time series can be obtained.

In general, there is overlap between successive sub-windows of the algorithm based on sliding window. So there are many kinds of conflicts when assigning the label of the sub-window to the sample points in the sub-window. For example, for adjacent sub-window A and B, the degree of overlap is 50. The prediction label obtained by sub-window A is label_A and the prediction label obtained by sub-window B is label B. When the sample points of the overlapping area of sub- window A and B are given by label, there are many kinds of conflicts, as shown in Fig. 1. Since all the algorithms based on sliding windows in this paper have a window overlap degree of 50, when the sub-window number starts from 0, we only retain the prediction label of the even number sub-window for predicting the sub-window label. The label is assigned to the sample point in the corresponding sub-window in turn. Finally, the label of each sample point in the whole time series is obtained.

### 5.4 Experiment evaluation

Because the HAR problem can be regarded as a classification problem, two evaluation indexes commonly used in classification problems, namely Accuracy (Acc) and F1-score (F1), are adopted.

The class of interest is called positive, and another is called negative. The positive which is predicted as positive is called True Positive (TP), the positive which is predicted as negative is called False Negative (FN), the negative which is predicted as positive is called False Positive (FP), and the negative which is predicted as negative is called True Negative (TN). Then the Accuracy is given by the formula (4), and F1-score is calculated by Precision (P) and Recall (R), which are given by formulas (5), (6) and (7) respectively.

$$Acc = \frac{TP+TN}{TP+FN+FP+TN} \quad (4)$$

$$P = \frac{TP}{TP+FP} \quad (5)$$

$$R = \frac{TP}{TP+FN} \quad (6)$$

$$\frac{2}{F1_i} = \frac{1}{P_i} + \frac{1}{R_i} \quad (7)$$

$F1_i$ is the F1-score of samples with label $i$, $P_i$ is the Precision of samples with label $i$, $R_i$ is the Recall of samples with label $i$. In fact, in addition to the Accuracy is calculated on all samples, the result of F1-score is actually a weighted F1-score. The calculation method is given by formula (8).

$$F1=\sum_i w_i \cdot F1_i \quad (8)$$

Where $F1$ is the weighted F1-score of all samples, $w_i$ is the proportion of samples with label $i$.

In order to ensure the relative objectivity of the experimental evaluation results, the traditional classification algorithm such as SVM、kNN、DT、QDA is from the open source machine learning algorithm library scikit-learn. The version used is 0.19.0. In the experiments, the CNN has a relatively simple structure, which contains convolution layer, pooling layer, full connection layering and soft-max layer. The FCN consists of a convolution layer and a pooling layer which are repeated six times in sequence. At the end of FCN, there are a $1\times1$ convolution layer, a deconvolution layer and a soft-max layer. The network structure of U-Net is described in detail in Section 4. And as mentioned above, for each algorithm, the parameters of the same algorithm are consistent on different datasets.

We summarize the performance of the seven algorithms on five datasets, and mark the highest score in bold, as shown in Table 1.

Table 1. The overall performance of seven algorithms on five datasets

| Dataset | Indicator | SVM | kNN | Decision Tree | QDA | CNN | FCN | U-Net |
|---|---|---|---|---|---|---|---|---|
| WISDM | Acc | 0.813 | 0.804 | 0.853 | 0.756 | 0.958 | 0.862 | **0.970** |
| | F1 | 0.848 | 0.813 | 0.850 | 0.779 | 0.958 | 0.861 | **0.970** |
| UCI HAR | Acc | 0.940 | 0.900 | 0.861 | 0.899 | 0.950 | 0.861 | **0.984** |
| | F1 | 0.941 | 0.901 | 0.862 | 0.901 | 0.950 | 0.862 | **0.984** |
| UCI HAPT | Acc | 0.918 | 0.887 | 0.807 | 0.854 | 0.917 | 0.929 | **0.931** |
| | F1 | 0.922 | 0.889 | 0.806 | 0.877 | 0.916 | 0.928 | **0.931** |
| Gesture | Acc | 0.785 | 0.710 | 0.700 | 0.798 | 0.914 | 0.910 | **0.947** |
| | F1 | 0.827 | 0.713 | 0.698 | 0.850 | 0.909 | 0.910 | **0.947** |
| Sanitation | Acc | 0.817 | 0.871 | 0.803 | 0.638 | 0.842 | 0.597 | **0.886** |
| | F1 | 0.820 | 0.871 | 0.804 | 0.646 | 0.839 | 0.592 | **0.885** |

On the whole, we can find that the U-Net method achieves higher scores than other algorithms in terms of Acc and F1. Especially for the Gesture dataset, the accuracy and F1-score of the U-Net method are up to 94.7%, 3.8% higher than the second highest score. The same result is obtained for the UCI HAR dataset, and the highest score of 98.4% in the five datasets is obtained.

CNN algorithm can also achieve better performance. For example, in WISDM dataset, the accuracy and F1-socre of U-Net method are almost the same. However, this is due to the fact that the 6 kinds of activities contained in the WISDM dataset are simple long term. In this case, the U-Net method may not have an absolute advantage. In addition, it can also be found that the performance of four traditional non-deep learning algorithms for Gesture datasets with a large number of short-term activities is not ideal, which makes them not suitable for recognition tasks with a large number of short-term activities. This also proves that U-Net method is not only suitable for simple long term activity recognition, but also more suitable for short term activity recognition. It has strong robustness.

It can be found that although FCN can also realize dense labelling, its performance is obviously worse than that of U-Net in our experiments. In most datasets, except on UCI HAPT dataset, the performance of FCN is also worse than that of CNN. What’s more, on Sanitation datasets, FCN obtains the worst results among all algorithms. Considering that the FCN architecture used in [35] is to add an up-sampling layer after the traditional CNN architecture, the low-resolution feature map produced by CNN is directly linearly expanded so that the output has the same size as the input. Unlike the U-Net method for dense prediction, the FCN method does not utilize high-resolution feature map information, but directly performs up-sampling operation, resulting in the loss of many shallow features which often contain position information. Theoretically, the FCN dense prediction results are rough, not as good as U-Net based on dense prediction, and even worse than traditional CNN based on sliding window prediction. This conclusion is consistent with the experimental results.

Remarkably, SVM and kNN also perform well on some datasets. For example, the performance of SVM is even better than that of CNN on UCI HAPT dataset. However, the performance of kNN on the Sanitation dataset is better than that of CNN, but the performance of SVM and CNN is not stable enough and the robustness is not high enough.

All the samples of each dataset are considered in the calculation of the above indexes. In fact, due to the heterogeneity among various types of samples, the class with the larger prop ortion of samples has greater influence on the overall result. It may be that on certain datasets, the algorithm scores high on a category with a large sample size, while scores lower on other categories, but the overall score remains high. We can't find the difference among different algorithms through the accuracy and weighted F1-score of all the samples in the dataset. Therefore, in order to show the differences of these algorithm more clearly, it is necessary to calculate the index of each algorithm on its different classes of samples for each dataset. For the sake of brevity and without losing generality. F1-score on different classes is selected as evaluation index only.

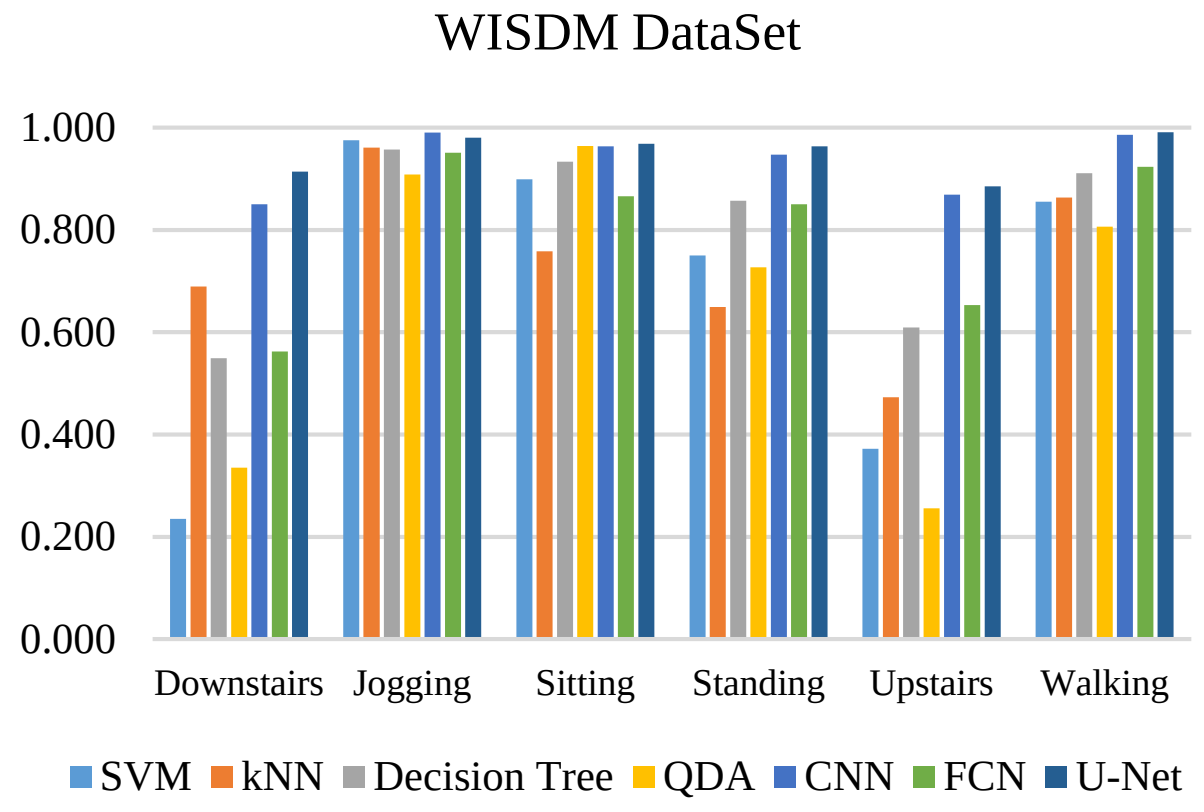

(a)

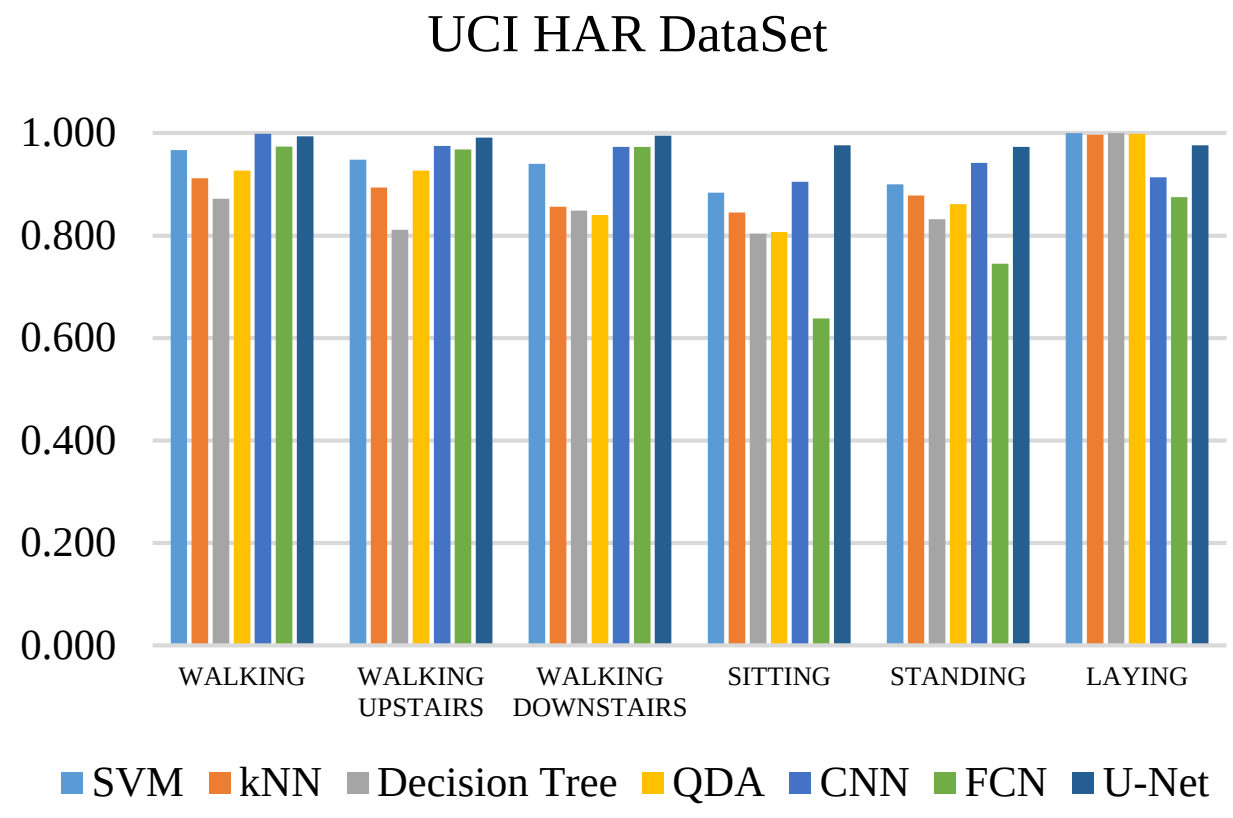

(b)

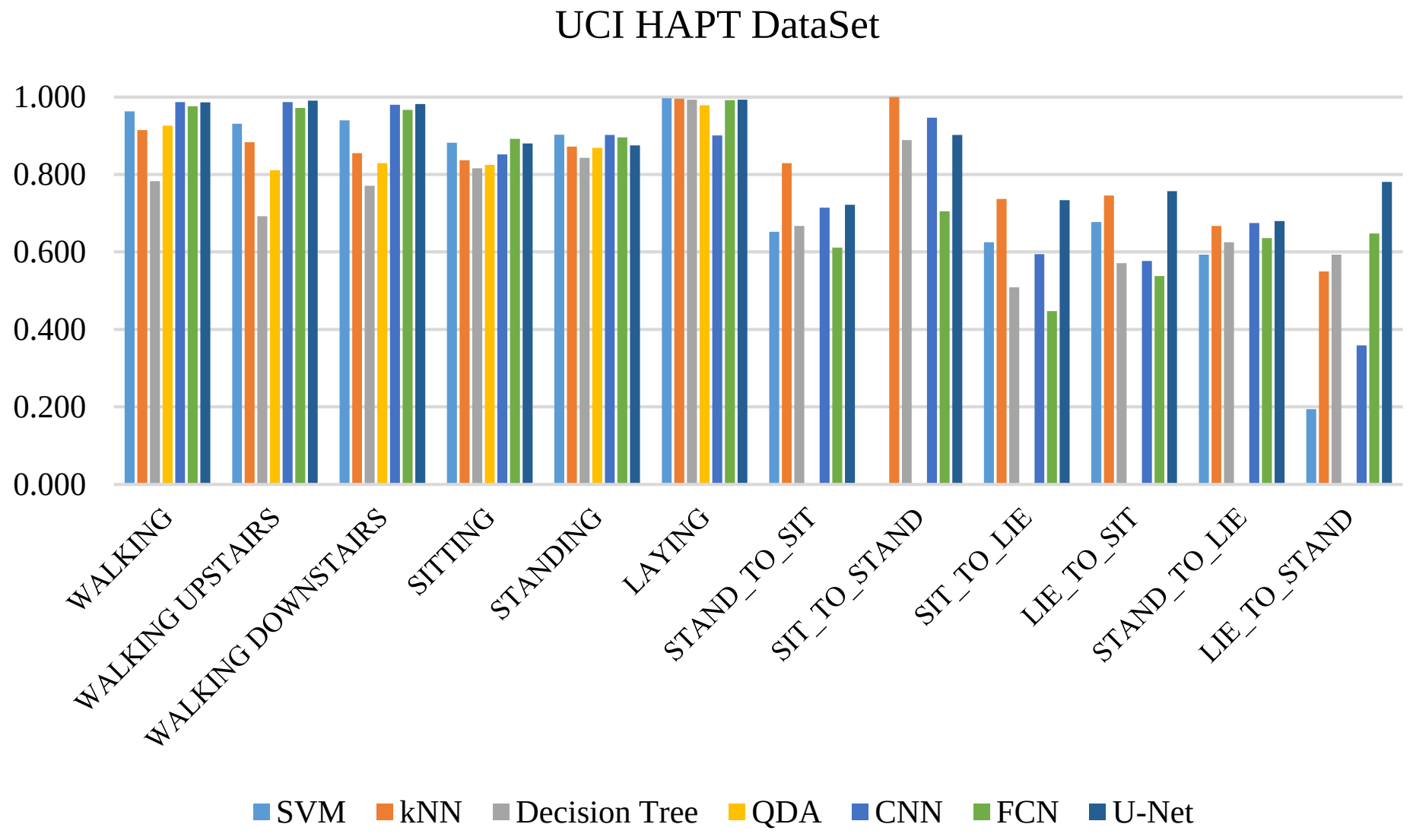

(c)

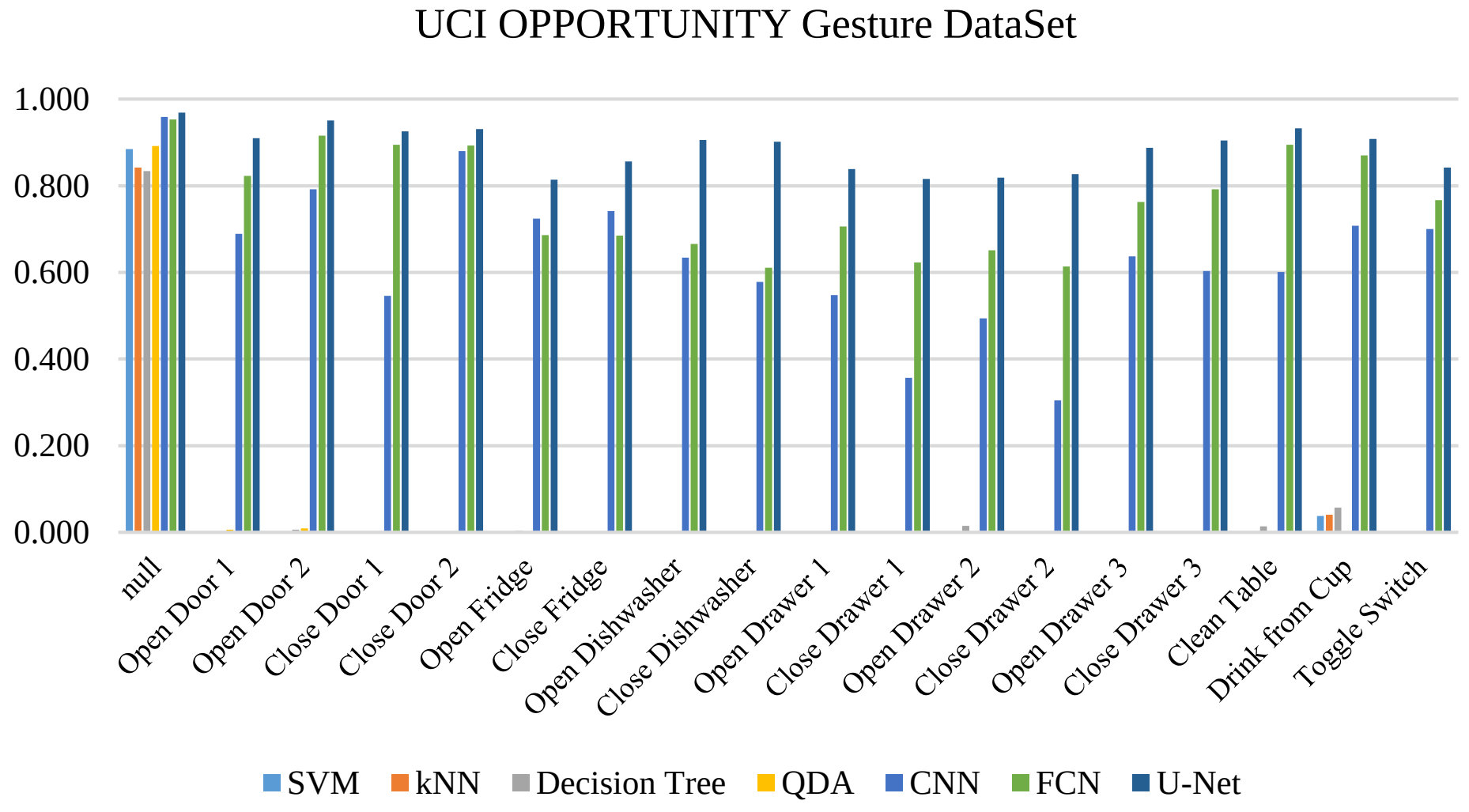

(d)

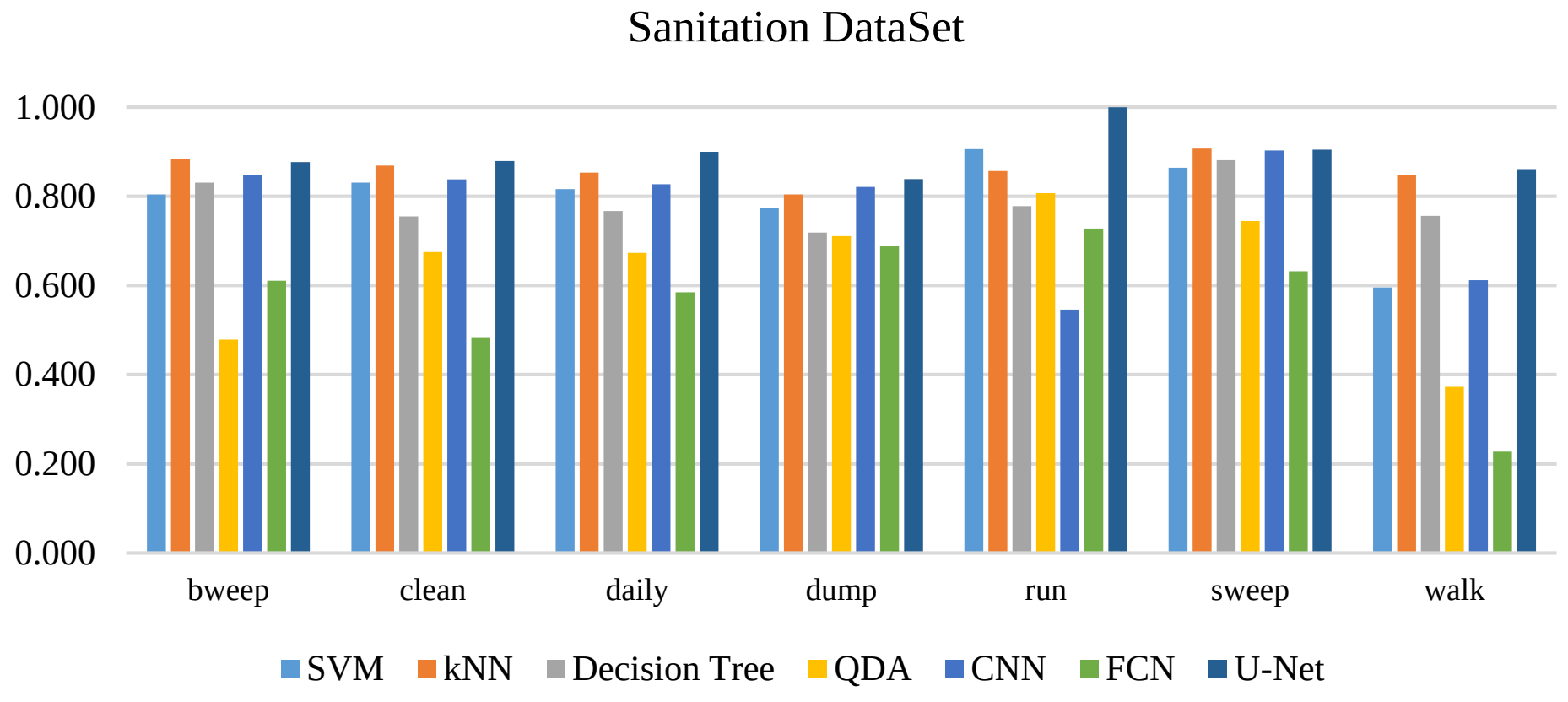

(e)

Figure 6. F1-score of seven algorithms on different samples of five datasets

Fig. 6 shows the seven algorithms in the experiment for F1-score on different class samples of five datasets. The horizontal coordinates of the histogram represent the labels of all kinds of samples. The ordinate represents F1-score. The seven bars of each label represent the F1-score of SVM, kNN, DT, QDA, CNN, FCN and U-Net on these samples. It can be seen that the performance of the proposed U-Net method is still better than other algorithms. Especially from Fig. 5 (d), we can find that the traditional algorithm can hardly recognize these short-term activities for UCI OPPORTRTUNITY dataset which contains a large number of short-term activities. As shown in Fig. 5(d), the first four algorithms have an F1-score of almost zero on these activities. Even for CNN, its performance is not satisfactory. And FCN, which can also achieve dense labelling, performs better but still has the potential for significant improvement. However, the U-Net method performs well in all kinds of activities.

The prediction time consumption of each algorithm on different datasets is used to evaluate the computational complexity. What really affects the prediction efficiency of the model is not the length of training time, but the actual prediction speed. Even if the model training time is short, as long as the prediction time is too long, it is not an efficient model. As shown in Table 2, it is the prediction time consumption of seven algorithm on five datasets. The time unit is in seconds (s), and each prediction dataset account for 30% of the total dataset.

Table 2. The prediction time consumption of seven algorithms on five datasets

| Dataset | SVM | kNN | Decision Tree | QDA | CNN | FCN | U-Net |
|---|---|---|---|---|---|---|---|
| WISDM | 0.179 | 0.343 | 0.000 | 0.004 | 0.055 | 0.128 | 0.508 |
| UCI HAR | 5.360 | 12.300 | 0.005 | 0.197 | 0.036 | 0.088 | 0.331 |
| UCI HAPT | 6.750 | 19.800 | 0.008 | 0.267 | 0.043 | 0.100 | 0.343 |
| Gesture | 53.200 | 237.000 | 0.008 | 0.265 | 0.096 | 0.162 | 0.426 |
| Sanitation | 0.158 | 0.186 | 0.000 | 0.005 | 0.018 | 0.062 | 0.129 |

It can be found that SVM and kNN take longer time on UCI HAR dataset, UCI HAPT dataset and UCI OPPORTUNITY Gesture dataset, indicating that the time efficiency of SVM and kNN is not very high. In addition, the other algorithms' prediction time is less than 1 second. Especially the DT’s prediction time is very short, because it only needs to perform branch judgment according to each feature. Since both CNN, FCN and U-Net use GPU to accelerate their predictions, which is not the same as the traditional algorithms that run on CPU, the prediction time consumption comparison between deep learning algorithms makes more sense. We find that the prediction time of CNN, FCN and U-Net increase in turn due to their different network layer number. However, the gap is very small, almost negligible. And U-Net has achieved a prediction speed of 0.1 seconds level on our machines, which can be further improved on better performance machines.

To sum up, whether the Accuracy or the weighted F1-score are considered as evaluation indicator, the proposed U-Net algorithm is obviously superior to other algorithms.

## 6 Conclusion

This paper presents a method of HAR based on time series using U-Net. Different from the traditional HAR method of using sliding window, our solution overcomes the problem of multi-class window inherent in the sliding window method, and realizes the classification labelling of each sample point in time series.

At the same time, it also effectively solves the problem that the existing methods can not recognize a large number of short-term activities. Moreover, the U-Net method is stable and robust in all test datasets. On the whole, the performance of the proposed U-Net method is superior to that of other algorithms.

## Acknowledgments

This work is supported by Fab. X Artificial Intelligence Research Center, Beijing, P.R.C.

## Reference

[1] Chen L, Hoey J, Nugent C D, et al. Sensor-based activity recognition[J]. IEEE Transactions on Systems, Man, and Cybernetics, Part C (Applications and Reviews), 2012, 42(6): 790-808.

[2] Kwapisz J R, Weiss G M, Moore S A. Activity recognition using cell phone accelerometers[J]. ACM SigKDD Explorations Newsletter, 2011, 12(2): 74-82.

[3] Brezmes T, Gorricho J L, Cotrina J. Activity recognition from accelerometer data on a mobile phone[J]. Distributed computing, artificial intelligence, bioinformatics, soft computing, and ambient assisted living, 2009: 796-799.

[4] Sun L, Zhang D, Li B, et al. Activity recognition on an accelerometer embedded mobile phone with varying positions and orientations[J]. Ubiquitous intelligence and computing, 2010: 548-562.

[5] Stephenson R M, Naik G R, Chai R. A system for accelerometer-based gesture classification using artificial neural networks[C]//Engineering in Medicine and Biology Society (EMBC), 2017 39th Annual International Conference of the IEEE. IEEE, 2017: 4187-4190.

[6] Panwar M, Dyuthi S R, Prakash K C, et al. CNN based approach for activity recognition using a wrist-worn accelerometer[C]//Engineering in Medicine and Biology Society (EMBC), 2017 39th Annual International Conference of the IEEE. IEEE, 2017: 2438-2441.

[7] Verma V K, Lin W Y, Lee M Y, et al. Levels of activity identification & sleep duration detection with a wrist-worn accelerometer-based device[C]//Engineering in Medicine and Biology Society (EMBC), 2017 39th Annual International Conference of the IEEE. IEEE, 2017: 2369-2372.

[8] Lee, M., Kim, J., Kim, K., Lee, I., Jee, S.H., and Yoo, S.K. 2009. Physical activity recognition using a single tri-axis accelerometer[C]//Proceedings of the world congress on engineering and computer science. 2009, 1.

[9] Gupta P, Dallas T. Feature selection and activity recognition system using a single triaxial accelerometer[J]. IEEE Transactions on Biomedical Engineering, 2014, 61(6): 1780-1786.

[10] Zhang M, Sawchuk A A. Motion primitive-based human activity recognition using a bag-of-features approach[C]//Proceedings of the 2nd ACM SIGHIT International Health Informatics Symposium. ACM, 2012: 631-640.

[11] Dong Z, Wejinya U C, Zhou S, et al. Real-time written-character recognition using MEMS motion sensors: Calibration and experimental results[C]//Robotics and Biomimetics, 2008. ROBIO 2008. IEEE International Conference on. IEEE, 2009: 687-691.

[12] Rezaie H, Ghassemian M. An Adaptive Algorithm to Improve Energy Efficiency in Wearable Activity Recognition Systems[J]. IEEE Sensors Journal, 2017, 17(16): 5315-5323.

[13] Karantonis D M, Narayanan M R, Mathie M, et al. Implementation of a real-time human movement classifier using a triaxial accelerometer for ambulatory monitoring[J]. IEEE transactions on information technology in biomedicine, 2006, 10(1): 156-167.

[14] Arif M, Bilal M, Kattan A, et al. Better physical activity classification using smartphone acceleration sensor[J]. Journal of medical systems, 2014, 38(9): 95.

[15] Jehad Sarkar, La The Vinh, Young-Koo Lee, et al. GPARS: a general-purpose activity recognition system[J]. Applied Intelligence, 2011, 35(2):242-259.

[16] Verma V K, Lin W Y, Lee M Y, et al. Levels of activity identification & sleep duration detection with a wrist-worn accelerometer-based device[C]//Engineering in Medicine and Biology Society (EMBC), 2017 39th Annual International Conference of the IEEE. IEEE, 2017: 2369-2372.

[17] González S, Sedano J, Villar J R, et al. Features and models for human activity recognition[J]. Neurocomputing, 2015, 167: 52-60.

[18] Van Hees V T, Gorzelniak L, Leon E C D, et al. Separating movement and gravity components in an acceleration signal and implications for the assessment of human daily physical activity[J]. PloS one, 2013, 8(4): e61691.

[19] Fan L, Wang Z, Wang H. Human activity recognition model based on Decision tree[C]//Advanced Cloud and Big Data (CBD), 2013 International Conference on. IEEE, 2013: 64-68.

[20] Xue Y, Jin L. A naturalistic 3D acceleration-based activity dataset & benchmark evaluations[C]//Systems Man and Cybernetics (SMC), 2010 IEEE International Conference on. IEEE, 2010: 4081-4085.

[21] He Z, Jin L. Activity recognition from acceleration data based on discrete consine transform and SVM[C]//Systems, Man and Cybernetics, 2009. SMC 2009. IEEE International Conference on. IEEE, 2009: 5041-5044.

[22] Hassan M M, Uddin M Z, Mohamed A, et al. A robust human activity recognition system using smartphone sensors and deep learning[J]. Future Generation Computer Systems, 2018, 81: 307-313.

[23] Reyes-Ortiz J L, Oneto L, Sama A, et al. Transition-aware human activity recognition using smartphones[J]. Neurocomputing, 2016, 171: 754-767.

[24] Mannini A, Intille S S, Rosenberger M, et al. Activity recognition using a single accelerometer placed at the wrist or ankle[J]. Medicine and science in sports and exercise, 2013, 45(11): 2193.

[25] Nguyen N D, Truong P H, Jeong G M. Daily wrist activity classification using a smart band[J]. Physiological Measurement, 2017, 38(9): L10-L16.

[26] Sun L, Zhang D, Li B, et al. Activity recognition on an accelerometer embedded mobile phone with varying positions and orientations[J]. Ubiquitous intelligence and computing, 2010: 548-562.

[27] Preece S J, Goulermas J Y, Kenney L P J, et al. A comparison of feature extraction methods for the classification of dynamic activities from accelerometer data[J]. IEEE Transactions on Biomedical Engineering, 2009, 56(3): 871-879.

[28] Long X, Yin B, Aarts R M. Single-accelerometer-based daily physical activity classification[C]//Engineering in Medicine and Biology Society, 2009. EMBC 2009. Annual International Conference of the IEEE. IEEE, 2009: 6107-6110.

[29] Lester J, Choudhury T, Kern N, et al. A hybrid discriminative/generative approach for modeling human activities[J]. 2005.

[30] Yang J. Toward physical activity diary: motion recognition using simple acceleration features with mobile phones[C]//Proceedings of the 1st international workshop on Interactive multimedia for consumer electronics. ACM, 2009: 1-10.

[31] Wang J, Chen Y, Hao S, et al. Deep learning for sensor-based activity recognition: A Survey[J]. Pattern Recognition Letters, 2018.

[32] Chen Y, Xue Y. A deep learning approach to human activity recognition based on single accelerometer[C]//Systems, Man, and Cybernetics (SMC), 2015 IEEE International Conference on. IEEE, 2015: 1488-1492.

[33] Preece S J, Goulermas J Y, Kenney L P J, et al. Activity identification using body-mounted sensors—a review of classification techniques[J]. Physiological measurement, 2009, 30(4): R1.

[34] Ordóñez F J, Roggen D. Deep Convolutional and LSTM Recurrent Neural Networks for Multimodal Wearable Activity Recognition[J]. Sensors, 2016, 16(1):115.

[35] Yao R, Lin G, Shi Q, et al. Efficient Dense Labelling of Human Activity Sequences from Wearables using Fully Convolutional Networks[J]. Pattern Recognition, 2017, 78.

[36] Zheng Y, Wong W K, Guan X, et al. Physical Activity Recognition from Accelerometer Data Using a Multi-Scale Ensemble Method[C]//IAAI. 2013.

[37] Long J, Shelhamer E, Darrell T. Fully convolutional networks for semantic segmentation[C]//Proceedings of the IEEE conference on computer vision and pattern recognition. 2015: 3431-3440.

[38] Ronneberger O, Fischer P, Brox T. U-net: Convolutional networks for biomedical image segmentation[C]//International Conference on Medical image computing and computer-assisted intervention. Springer, Cham, 2015: 234-241.

[39] Badrinarayanan V, Kendall A, Cipolla R. Segnet: A deep convolutional encoder-decoder architecture for image segmentation[J]. IEEE transactions on pattern analysis and machine intelligence, 2017, 39(12): 2481-2495.

[40] Noh H, Hong S, Han B. Learning deconvolution network for semantic segmentation[C]//Proceedings of the IEEE International Conference on Computer Vision. 2015: 1520-1528.

[41] Jansson A, Humphrey E, Montecchio N, et al. Singing voice separation with deep U-Net convolutional networks[J]. 2017.

[42] Ng A Y, Jordan M I. On discriminative vs. generative classifiers: A comparison of logistic regression and naive bayes[C]//Advances in neural information processing systems. 2002: 841-848.

[43] He Z Y, Jin L W. Activity recognition from acceleration data using AR model representation and SVM[C]//Machine Learning and Cybernetics, 2008 International Conference on. IEEE, 2008, 4: 2245-2250.

[44] Khan A M, Lee Y K, Lee S Y, et al. A triaxial accelerometer-based physical-activity recognition via augmented-signal features and a hierarchical recognizer[J]. IEEE transactions on information technology in biomedicine, 2010, 14(5): 1166-1172.

[45] Lee S, Le H X, Ngo H Q, et al. Semi-Markov conditional random fields for accelerometer-based activity recognition[J]. Applied Intelligence, 2011, 35(2): 226-241.

[46] Lee S M, Sang M Y, Cho H. Human activity recognition from accelerometer data using Convolutional Neural Network[C]// IEEE International Conference on Big Data and Smart Computing. IEEE, 2017:131-134.

[47] Guan Y, Plötz T. Ensembles of deep lstm learners for activity recognition using wearables[J]. Proceedings of the ACM on Interactive, Mobile, Wearable and Ubiquitous Technologies, 2017, 1(2): 11.

[48] Edel M, Köppe E. Binarized-blstm-rnn based human activity recognition[C]//Indoor Positioning and Indoor Navigation (IPIN), 2016 International Conference on. IEEE, 2016: 1-7.

[49] Hammerla N Y, Halloran S, Ploetz T. Deep, Convolutional, and Recurrent Models for Human Activity Recognition using Wearables[J]. Journal of Scientific Computing, 2016, 61(2):454-476.

[50] Mittelman R. Time-series modeling with undecimated fully convolutional neural networks[J]. Computer Science, 2015.

[51] Kingma D P, Ba J. Adam: A method for stochastic optimization[J]. arXiv preprint arXiv:1412.6980, 2014.

[52] Anguita D, Ghio A, Oneto L, et al. A Public Domain Dataset for Human Activity Recognition using Smartphones[C]//ESANN. 2013.

[53] Chavarriaga R, Sagha H, Calatroni A, et al. The Opportunity challenge: A benchmark database for on-body sensor-based activity recognition[J]. Pattern Recognition Letters, 2013, 34(15): 2033-2042.